\documentclass{article}

\usepackage{arxiv}

\usepackage[utf8]{inputenc} 
\usepackage[T1]{fontenc}    
\usepackage{hyperref}       
\usepackage{url}            
\usepackage{booktabs}       
\usepackage{amsfonts}       
\usepackage{amsmath}
\usepackage{nicefrac}       
\usepackage{microtype}      
\usepackage{cleveref}       
\usepackage{lipsum}         
\usepackage{graphicx}
\usepackage{natbib}
\usepackage{doi}
\usepackage{graphicx}
\usepackage{subcaption}
\usepackage{array}
\usepackage{csquotes}
\usepackage{tabularx}
\usepackage{listings}
\usepackage[table]{xcolor}

\providecommand{\Description}[1]{}

\title{Do Vision-Language-Action Models Understand Instructions?\\A Mechanistic Interpretability Study on Language Grounding}

\newif\ifuniqueAffiliation
\uniqueAffiliationtrue

\ifuniqueAffiliation 
\author{ Theodor Wulff \\
	Department of Computer Science\\
	The University of Manchester\\
	Manchester, United Kingdom \\
	\texttt{theodor.wulff@manchester.ac.uk} \\
	\And
	Angelo Cangelosi \\
	Department of Computer Science\\
	The University of Manchester\\
	Manchester, United Kingdom \\
	\texttt{angelo.cangelosi@manchester.ac.uk} \\
}
\else
\usepackage{authblk}

\newbox{\orcid}\sbox{\orcid}{\includegraphics[scale=0.06]{orcid.pdf}} 
\author[1]{%
	\href{https://orcid.org/0000-0000-0000-0000}{\usebox{\orcid}\hspace{1mm}David S.~Hippocampus\thanks{\texttt{hippo@cs.cranberry-lemon.edu}}}%
}
\author[1,2]{%
	\href{https://orcid.org/0000-0000-0000-0000}{\usebox{\orcid}\hspace{1mm}Elias D.~Striatum\thanks{\texttt{stariate@ee.mount-sheikh.edu}}}%
}
\affil[1]{Department of Computer Science, Cranberry-Lemon University, Pittsburgh, PA 15213}
\affil[2]{Department of Electrical Engineering, Mount-Sheikh University, Santa Narimana, Levand}
\fi

\renewcommand{\headeright}{Preprint}
\renewcommand{\undertitle}{Preprint}
\renewcommand{\shorttitle}{Do VLAs Understand Instructions?}

\begin{document}

\maketitle 


\begin{abstract}
    Vision-Language-Action models are designed to generalise across environments and task descriptions, raising the question of whether their action generation actually depends on the language instruction, or whether they largely rely on visual cues and superficial correlations.
    Robustness to variance in the visual and linguistic observation space is critical for real-world deployment, yet VLAs lack explicit grounding modules and instead rely on the intrinsic language grounding capabilities of their Vision-Language model backbones.
    For this reason, we conduct a controlled mechanistic interpretability study on the language grounding capabilities of two state-of-the-art Vision-Language-Action models, $\pi_{0.5}$ and GR00T N1.7, by applying activation and attribution patching to the residual stream of the action generation modules.
    We systematically corrupt the task instruction of input samples of the LIBERO benchmark following five strategies: synonym replacement, semantic scaling, directional corruption, random object substitution, and empty string.
    Our experiments find that both models are comparatively insensitive to abstract rephrasing and to referencing non-existent objects, but react strongly to empty task descriptions and, especially, to directional language.
    During action generation, this sensitivity is concentrated in different loci for each model: mainly in the early, periodic cross-attention layers for GR00T N1.7, versus distributed across the earliest and selected later layers for $\pi_{0.5}$. For GR00T N1.7, directional perturbations drive some of the largest causal effects while leaving the internal representational geometry comparatively unchanged, a dissociation we do not observe clearly for $\pi_{0.5}$. Finally, the reliability of attribution patching is model-dependent: it closely tracks activation patching for GR00T N1.7 but not for $\pi_{0.5}$.
\end{abstract}

\section{Introduction}
Robotic systems deployed in real-world environments must be robust, as unpredictable behaviour may lead to severe and potentially catastrophic consequences.
Vision-Language-Action models (VLAs) achieve impressive generalisation across different environments and task formulations, and yet the internal mechanisms underlying this behaviour remain largely a black box.
This creates a gap between empirical performance and scientific understanding.
Mechanistic interpretability has yielded insights in Large Language~\cite{wangInterpretabilityWildCircuit2022} and Vision models~\cite{gohMultimodalNeuronsArtificial2021}, but only a handful of works exist that investigate Vision-Language-Action models from a model-internal perspective~\cite{grantNotAllFeatures2026,miaoContrastiveConceptorActivation2026}, particularly with respect to their action decoding strategies.
While findings in NLP and Computer Vision might transfer to the shared underlying Language and Vision-Language model backbones, VLA action generation strategies regularly introduce specific action expert modules with varying decoding strategies that particularly require the generation of language-conditioned outputs.

To this end, we address the gap between generalisability and understanding of the model's internal computations by analysing the action expert modules of two state-of-the-art VLAs: $\pi_{0.5}$~\cite{black$pi_05$VisionLanguageActionModel2025} and GR00T N1.7~\cite{GR00TN1Open2025}.
Like many VLAs, both models employ a modular architecture in which an action expert incorporates visuolinguistic information from a Vision-Language model (VLM) backbone into a conditional flow-matching process for action generation.
However, they differ in how this conditioning is implemented: $\pi_{0.5}$ integrates language via shared attention between action expert and backbone layers, while GR00T uses dedicated masked cross-attention layers.
This contrast lets us ask whether such differences in conditioning have measurable consequences for how language is causally used inside the action expert.

We apply the mechanistic interpretability techniques attribution patching and activation patching to the residual stream of the transformer layers in each model's action expert, and complement these causal measures with Centred Kernel Alignment (CKA) between clean and perturbed activations.
Pairing a causal measure with a representational-similarity measure lets us distinguish two failure modes that are otherwise easy to conflate: (i) where a model's representations are simply insensitive to a perturbation, and (ii) where representations shift but the action expert routes only a subset of that shift into behaviourally relevant computation.
By following a systematic language perturbation scheme, we construct the required corrupted instructions from the clean instructions for the corresponding patching methods.
Our perturbation scheme operates on five categories of increasing severity: synonym replacements, semantic scaling, directional perturbations, random object substitution, and empty string.
While the first two address sensitivity toward less specific inputs and alternate formulations on a verb and noun level, the directional perturbations directly investigate the effect of processing directional attributes within the action expert. 
We test whether the model can fall back on purely visual information by rendering the task impossible due to referencing objects not visible in the current scene or dropping the instruction altogether.
Since all perturbed predictions are evaluated against the original demonstration, our experiments establish whether and where language causally influences the action, not whether it is interpreted correctly.

Specifically, our contributions can be summarised as follows:
\begin{itemize}
    \item \textbf{A combined causal/representational analysis framework for diffusion/flow-matching VLA action experts.}
    We integrate attribution patching, activation-patching verification, and CKA into a single evaluation framework applicable to transformer-based diffusion action heads, and use activation patching to directly quantify where attribution patching's linear approximation deviates from the true causal effect of language conditioning, and for which model this approximation is reliable.
    
    \item \textbf{Language-perturbation sensitivity is model-specific.} 
    GR00T N1.7 exhibits a periodic layer-wise pattern that coincides with the placement of its cross-attention conditioning blocks, whereas $\pi_{0.5}$ shows stronger effects concentrated in early and selected later layers. 
    Because the two models differ in more than their conditioning mechanism, we regard the link to architecture as a hypothesis.
    $\pi_{0.5}$ is comparatively more sensitive to verb perturbations, while GR00T N1.7 is more sensitive to noun perturbations. 
    GR00T N1.7 also exhibits larger CKA shifts across several perturbation conditions, whereas $\pi_{0.5}$'s global representational geometry remains relatively stable.
    
    \item \textbf{Causal attribution and representational geometry dissociate for directional language in GR00T N1.7.} 
    In GR00T N1.7, directional perturbations elicit some of the largest attribution-patching effects of any perturbation type, yet leave the residual stream's representational geometry, as measured by CKA, comparatively unchanged. 
    We do not observe this dissociation clearly for $\pi_{0.5}$.
    For synonym, hypernym, and empty-string perturbations, causal and representational effects track one another more closely and follow our proposed perturbation-severity ranking. 
    We hypothesise that this dissociation arises because directional phrases occupy a narrow, tightly clustered region of the embedding space, so the model's action output is highly sensitive to the perturbation despite the underlying representation shifting comparatively little.
\end{itemize}

\begin{figure*}
    \centering
    \includegraphics[width=\linewidth]{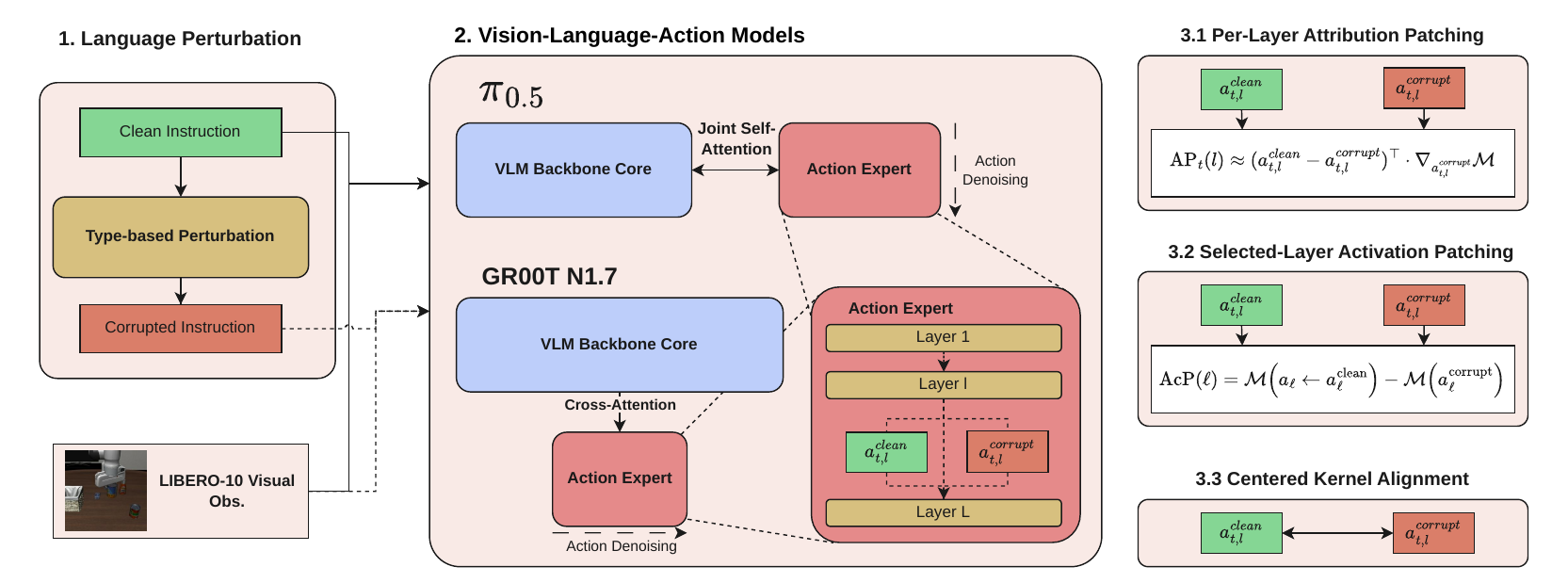}
    \caption{\textbf{Method Overview.} We study how $\pi_{0.5}$ and GR00T N1.7 process linguistic instructions during action decoding. (1) We corrupt the clean instructions following six different perturbation strategies. (2) We record the residual stream activations of both models during action decoding. (3) We analyse the causal and representational changes using attribution patching (3.1), activation patching (3.2), and Centered Kernel Alignment (3.3)}
    \Description{Method Overview. We study how $\pi_{0.5}$ and GR00T N1.7 process linguistic instructions during action decoding. (1) We corrupt the clean instructions following six different perturbation strategies. (2) We record the residual stream activations of both models during action decoding. (3) We analyse the causal and representational changes using attribution patching (3.1), activation patching (3.2), and Centered Kernel Alignment (3.3)}
    \label{fig:method_overview}
\end{figure*}
\section{Related Work}
\subsection{Vision-Language-Action Models}
Inspired by the successes and generalisable capabilities of foundation models in domains such as NLP and image generation~\cite{openaiGPT4TechnicalReport2023,dubeyLlama3Herd2024,betkerImprovingImageGeneration,rombachHighResolutionImageSynthesis2022}, Vision-Language-Action models aim to create robot policies that generalise across tasks, environments and embodiments~\cite{liuRDT1BDiffusionFoundation2024,teamGeminiRoboticsBringing2025,SpatialVLA2025,belkhaleRTHActionHierarchies2024,GR00TN1Open2025,kimOpenVLAOpenSourceVisionLanguageAction2024a}. 
VLAs are usually built on top of pretrained VLM backbones to inherit their vision-language understanding. 
They either add a dedicated action decoder module~\cite{black$pi_05$VisionLanguageActionModel2025,GR00TN1Open2025} to generate embodiment-specific trajectories or utilize the VLM vocabulary to assign discrete actions to tokens~\cite{kimOpenVLAOpenSourceVisionLanguageAction2024a}. 
The action decoding strategies employed by VLAs range from simple linear mappings to sophisticated diffusion-based action decoders. 
The latter employ flow matching or traditional diffusion denoising strategies as used by Chi et al.~\citep{chiDiffusionPolicyVisuomotor2023}. 
As input, they take at least a task instruction in natural language alongside a visual observation.
While VLAs show promising directions to generalising to differently formulated instructions as seen by their task success rates, it remains unclear how strong their language grounding actually is. 

\subsection{Mechanistic Interpretability for Multimodal Foundation Models}
Early mechanistic interpretability work at the intersection of vision and language focused on foundation models, uncovering multimodal neurons that respond to shared visual-linguistic concepts \cite{gohMultimodalNeuronsArtificial2021} and interpreting CLIP's image representations directly \cite{gandelsmanInterpretingCLIPsImage2023,radfordLearningTransferableVisual2021a}. Sparse autoencoders have since been widely applied to further disentangle CLIP's internal representations \cite{raoDiscoverthenNameTaskAgnosticConcept2024,roSemanticsDisentanglingInformation2026,josephSteeringCLIPsVision2025}, while causal tracing tools originally developed for LLMs have been extended to vision-language models \cite{palitVisionLanguageMechanisticInterpretability2023a,basuUnderstandingInformationStorage2024,neoInterpretingVisualInformation2024a} and, more recently, video models \cite{shiCausalityMattersHow2026}, typically to localize task-relevant subgraphs \cite{rajaramAutomaticDiscoveryVisual} or via probing-based analysis \cite{salinAreVisionLanguageTransformers2022a}. 
Extending this line of work to embodied settings, recent studies have begun steering vision-language-action models (VLAs) \cite{haonMechanisticInterpretabilitySteering2025}, with a concurrent mechanistic study across five VLAs finding that these models process heterogeneous token sequences interleaving vision, language, and proprioception, and that mean-pooling activations across token positions destroys action-critical information despite high reconstruction quality \cite{grantNotAllFeatures2026}. 
To the best of our knowledge, no prior study has systematically examined, at the level of internal model computation, how VLAs respond to different types of perturbations in the language space, or where this processing occurs within their action-decoding components.
We address this gap by analysing the token-generation process in the action expert of $\pi_{0.5}$ and the action decoder of GR00T N1.7 using attribution and activation patching alongside representational analyses.

\subsection{Circuit Discovery}
Mechanistic interpretability treats neural networks as computational graphs, with nodes corresponding to attention heads, MLPs, or individual neurons and features, and edges representing activation flow between them~\cite{heimersheimHowUseInterpret2024}, a framing established by foundational work on transformer circuits~\cite{MathematicalFrameworkTransformer}, circuit-level case studies such as the IOI circuit~\cite{wangInterpretabilityWildCircuit2022} and the greater-than arithmetic circuit~\cite{hannaHowDoesGPT22023a}, and formalized theoretically through causal abstraction~\cite{geigerCausalAbstractionTheoretical2025a}. 
Early circuit discovery relied on manual activation patching, that was later refined into path patching and automated via ACDC, which systematically searches a model's computational graph for task-relevant subgraphs~\cite{conmyAutomatedCircuitDiscovery2023}, alongside related causal techniques such as causal mediation analysis, interchange interventions, and causal tracing~\cite{basuUnderstandingInformationStorage2024}. 
Because these intervention-based methods scale poorly, attribution patching emerged as a more efficient alternative: 
Attribution Patching approximates activation-patching effects via gradients~\cite{AttributionPatchingActivation,kramarAtPEfficientScalable2024}, and subsequent work has shown attribution patching can outperform automated circuit discovery empirically~\cite{syedAttributionPatchingOutperforms2024}. 
More recently, sparse autoencoders have enabled higher-resolution circuit analysis by exposing individual features as causally implicated nodes~\cite{marksSparseFeatureCircuits2025}, though at the cost of vastly larger node counts that make direct intervention prohibitive; this has motivated frontier approaches combining cross-layer transcoders with backward Jacobian tracing to construct partial feature-level circuit maps~\cite{lindseyBiologyLargeLanguage}, including per-prompt attribution graphs over transcoder features in production-scale models.
\section{Background}
\label{sec:background}
 
\subsection{Vision-Language-Action Model Architecture}
Vision-Language-Action models comprise a Vision-Language model core to process the incoming observation consisting of an instruction in natural language and camera inputs.
Specifically, we investigate the world understanding ability of the $\pi_{0.5}$ and the GR00T N1.7 VLAs through the lens of language grounding. 
Both models consist of a VLM backbone and a specific action expert module. 
The $\pi_{0.5}$ backbone is realized as SigLIP (400M)~\cite{zhaiSigmoidLossLanguage2023} and Gemma (2.6B)~\cite{teamGemma3Technical2025}, while the action expert is itself a 300M parameter transformer module. 
The action expert shares the transformer stack with the backbone VLM and interacts with it via block-wise, causally masked joint self-attention.
On the other hand, GR00T N1.7 uses a Cosmos-Reason2-2B VLM\footnote{\url{https://www.jetson-ai-lab.com/models/cosmos-reason2-2b/}}~\cite{nvidiaCosmosReason1PhysicalCommon2025} backbone to encode the observation into visuolinguistic embeddings, which are infused into the action generation module via cross-attention.
The action generation module is a transformer-based diffusion model with 32 alternating cross- and self-attention layers. 

Both action generation modules are diffusion-transformer architectures trained with a conditional flow-matching objective. 
Given a ground-truth action chunk $x_1$ and a noise sample $x_0 \sim \mathcal{N}(0, I)$, a noisy intermediate action representation is constructed via linear interpolation
\begin{equation}
    x_t = (1-t)\,x_0 + t\,x_1, \qquad t \sim \mathcal{U}[0,1],
\end{equation}
with corresponding target velocity $v_t^{\star} = x_1 - x_0$. 
The action expert is trained to predict this velocity conditioned on the current noisy action, the timestep, and the visuolinguistic embeddings produced by the VLM backbone, $o$:
\begin{equation}
    \mathcal{L}_{\text{FM}}(\theta) = \mathbb{E}_{t,\, x_0,\, x_1}\left[\; \big\| v_\theta(x_t, t, o) - (x_1 - x_0) \big\|^2 \;\right].
\end{equation}
Crucially, the two architectures diverge on the conditioning term $o$: 
GR00T N1.7 injects $o$ via alternating cross-attention layers, while the action expert of $\pi_{0.5}$ attends to $o$ through shared, block-wise causally-masked self-attention described above.
GR00T N1.7 additionally applies a masking scheme to its cross-attention layers, so that the model only cross-attends to linguistic tokens every $n$-th layer (by default,$n=4$).

Because the action expert has access to the observation exclusively through the conditioning embeddings $o$, the grounding quality is entirely inherited from the VLM backbone. 
Neither model employs a dedicated grounding module at inference time; hence, grounding instead emerges implicitly from the visuolinguistic representations the backbone acquired during pretraining and any subsequent VLA fine-tuning. 
This makes the backbone's world understanding a direct bottleneck on downstream action quality, motivating our investigation into how sensitive the action outputs of $\pi_{0.5}$ and GR00T N1.7 are to controlled perturbations of the language instruction.

\subsection{Activation Patching}

Activation patching~\cite{mengLocatingEditingFactual2022} measures the causal importance of a model component by directly intervening on its activations during a forward pass. 
Given a \textit{clean} input and a \textit{corrupted} input that differ in a targeted way, a forward pass is first run on each input, and the resulting activations are cached. 
A second forward pass is then run on the corrupted input, except that at a chosen layer $\ell$ and position, the cached corrupted activation $a^\text{corrupt}_\ell$ is replaced with the corresponding clean activation $a^\text{clean}_\ell$:
\begin{equation}
    \text{AcP}(\ell) = \mathcal{M}\big(a_\ell \leftarrow a^\text{clean}_\ell\big)
    - \mathcal{M}\big(a^\text{corrupt}_\ell\big),
\end{equation}
where $\mathcal{M}$ is a scalar metric evaluated on the resulting output, and $\mathcal{M}(a_\ell \leftarrow a^\text{clean}_\ell)$ denotes the metric obtained after substituting the clean activation at layer $\ell$ into the
otherwise-corrupted run. 
A large change in $\mathcal{M}$ indicates that the patched component carries information causally relevant to the behaviour under study.
In our case $\mathcal{M}$ is the standard flow matching L2 loss between the target velocities $v_t^{\star}$ and the predicted velocities $v_t^{corrupt}$ of the corrupted run.

The downside of activation patching is that this substitution must be performed separately for each layer and position of interest, requiring a full additional forward pass per measurement. 
For large models with many components, this makes exhaustive, fine-grained patching computationally expensive.

\subsection{Attribution Patching}
Attribution patching~\cite{AttributionPatchingActivation} reduces the computational cost of activation patching by approximating the effect of a change in input using two forward passes and a single backward pass.
It then approximates the causal effect of an activation $a$ at layer $\ell$ using a first-order Taylor expansion involving the gradients of the corrupted run:
\begin{equation}
    \text{AP}(\ell) \approx (a^\text{clean}_\ell - a^\text{corrupt}_\ell)^\top \cdot
    \nabla_{a^\text{corrupt}_\ell} \mathcal{M}
\end{equation}
This requires only two forward passes and one backward pass, making it tractable for the VLAs investigated in this study.
Attribution patching has the tendency to over- or underestimate the importance of different model layers, which is why we run additional activation patching trials on the layers identified using the results of the attribution patching runs. 



\section{Method}
We investigate the inner workings of Vision-Language-Action models using mechanistic interpretability techniques applied to the action expert modules of the VLAs $\pi_{0.5}$ and GR00T N1.7. 
Since the action expert's only access to the observation is through the VLM backbone's conditioning embeddings, any weaknesses in the backbone's language grounding should be directly observable in how the action expert responds to controlled changes in the language input. 
We therefore focus on the influence of systematically perturbed language instructions, using them as a probe for the robustness of the model's language understanding as it propagates through to action generation. 
Using attribution patching and verifying the resulting candidate components with exact activation patching, we identify which layers of the VLA action expert are particularly reactive to changes in the language domain, thereby localising where and how strongly linguistic grounding influences downstream action prediction.
\subsection{Dataset}
\label{sec:dataset}
We perform all our experiments using data extracted from the LIBERO benchmark~\cite{liuLIBEROBenchmarkingKnowledge2023}, as it is a common benchmark used for training and evaluating VLAs.
Specifically, we use a publicly available subset of offline demonstrations, LIBERO-10\footnote{https://huggingface.co/datasets/lerobot/libero\_10\_image}, which we select for its long-horizon, compositional task instructions.
The dataset contains 378 episodes, accumulating approximately 101,000 steps across 10 different tasks.  
From this pool, we randomly sample a subset of 10,000 steps.
For each sampled step, we use both available camera views (a wrist camera and a table-view camera) together with the robot's vectorized proprioceptive state and the task instruction as model inputs.

\subsection{Type-based Perturbation}
We generate the corrupt samples required for attribution patching by applying different perturbations exclusively to the language input, e.g., the task instruction.
When a perturbation is not applicable to a given instruction, we exclude the corresponding sample from the corresponding experiment. 
This occurs, for example, when Directional Corruption is applied to an instruction containing no directional term.
In principle, both Synonym Replacement and Semantic Scaling can be applied to almost any word in an instruction. 
In practice, we target noun and verb phrases specifically, as these most directly encode what should be done and which object should be interacted with.
We define five perturbation types, ranging from minimal single-word substitutions to complete removal of the instruction, to probe language robustness at increasing levels of severity:

\paragraph{Synonym Replacement.}
This perturbation investigates the robustness of the semantic representations in the chosen VLAs.
We expect replacing one synonym with another to have little to no effect on the outcome, especially in the later layers.
In practice, a target word is replaced with a semantically equivalent synonym.
Candidate synonyms for each verb and noun in the LIBERO-10 instructions were generated using Claude Opus 4.6 and verified against ConceptNet~\cite{speerConceptNet55Open2018} and WordNet~\cite{millerWordNetLexicalDatabase1994} to ensure semantic validity.

\paragraph{Semantic Scaling.}
This perturbation investigates the capacity for abstract language understanding of the VLAs by replacing target words with more abstract hypernyms.
For instance, \enquote{apple} would become \enquote{fruit}.
As with Synonym Replacement, candidate hypernyms were generated using Claude Opus 4.6 and verified against ConceptNet and WordNet.
Since hypernym substitution reduces referential specificity, we expect an effect stronger than under Synonym Replacement but weaker than under Random Object Replacement.
This tests whether model performance degrades gracefully as abstraction increases.

\paragraph{Directional Corruption.}
This perturbation investigates where the model processes directional information. 
To this end, we invert the directions given in the input prompt. 
For instance, \enquote{up} becomes \enquote{down} and \enquote{left} becomes \enquote{right}. 
Note that this excludes directional particles in phrasal verbs such as \enquote{pick up}. 
Instead, only stand-alone directional modifiers, as in \enquote{the \textit{left} plate}, are inverted.
We expect this to produce a large, clearly visible effect in the models' layer attributions.
This effect should be visible starting from the early layers, as directional information directly determines the trajectory of the output action.

\paragraph{Random Object Replacement.}
We generate a list of 50 object words using Claude Opus 4.6, none of which are present in any LIBERO scene.
For each noun in the task description, we randomly select one of these objects to replace the original.
This tests model behaviour when the instruction references an object absent from the scene, rendering the task unsolvable by design.
Accordingly, we expect this to produce a large effect in the layer attributions of the VLAs, as the instruction now references information absent from the visual observation.

\paragraph{Empty String.}
This perturbation replaces the task instruction entirely with an empty string.
As the most drastic manipulation, this condition mirrors the counterfactual null prompt of~\cite{grantNotAllFeatures2026}, and lets us test whether the model falls back on purely visual grounding in the absence of language input, revealing where language processing occurs during action generation.

\subsection{Attribution and activation patching}
For both $\pi_{0.5}$ and GR00T N1.7, we intervene on the residual stream of each transformer layer in the action expert, immediately after the output projection (\texttt{o\_proj}) of the self-attention block. 
This intervention point allows us to isolate the causal contribution of each layer's self-attention output to the downstream action prediction, independent of the subsequent MLP block. 
As the language conditioning in both models enters the action expert through attention (via cross-attention for GR00T N1.7 and joint self-attention for $\pi_{0.5}$), patching directly after \texttt{o\_proj} lets us localize, layer by layer, where perturbed language information first begins to diverge from the clean run.

The reported effect at each layer and token corresponds to the attribution patching approximation of the change in $\mathcal{L}_{\text{FM}}$ induced by substituting the corrupted activation at that location. 
We report attribution patching effects per layer and token as heatmaps. 
Additionally, we report Centered Kernel Alignment (CKA)~\cite{kornblithSimilarityNeuralNetwork2019} plots per layer and perturbation type. 
Given two representation matrices over the same set of inputs, CKA measures the similarity of their pairwise similarity structures after centering.

\section{Experimental Setup}
\label{sec:experimental_setup}
To ensure we investigate strong-performing VLAs, we use LIBERO-10 fine-tuned checkpoints for both models. For $\pi_{0.5}$, we use the checkpoint publicly available on HuggingFace\footnote{\url{https://huggingface.co/lerobot/pi05_libero_finetuned_v044}}. 
For GR00T N1.7, we similarly use a LIBERO-10 fine-tuned checkpoint, also available on HuggingFace\footnote{\url{https://huggingface.co/nvidia/gr00t17-lerobot-libero_10-640}}.
We extract clean activations, corrupted activations, and gradients using 10,000 samples from the LIBERO-10 offline dataset subset described previously. 
All experiments are run on a single NVIDIA A100 GPU, using each checkpoint at its default precision as released, which may vary by component (i.e., mixed precision).
Results are aggregated across the randomly selected sample population.

\begin{figure*}[htb]
    \centering
    \begin{subfigure}[b]{0.5\textwidth}
        \includegraphics[width=\linewidth]{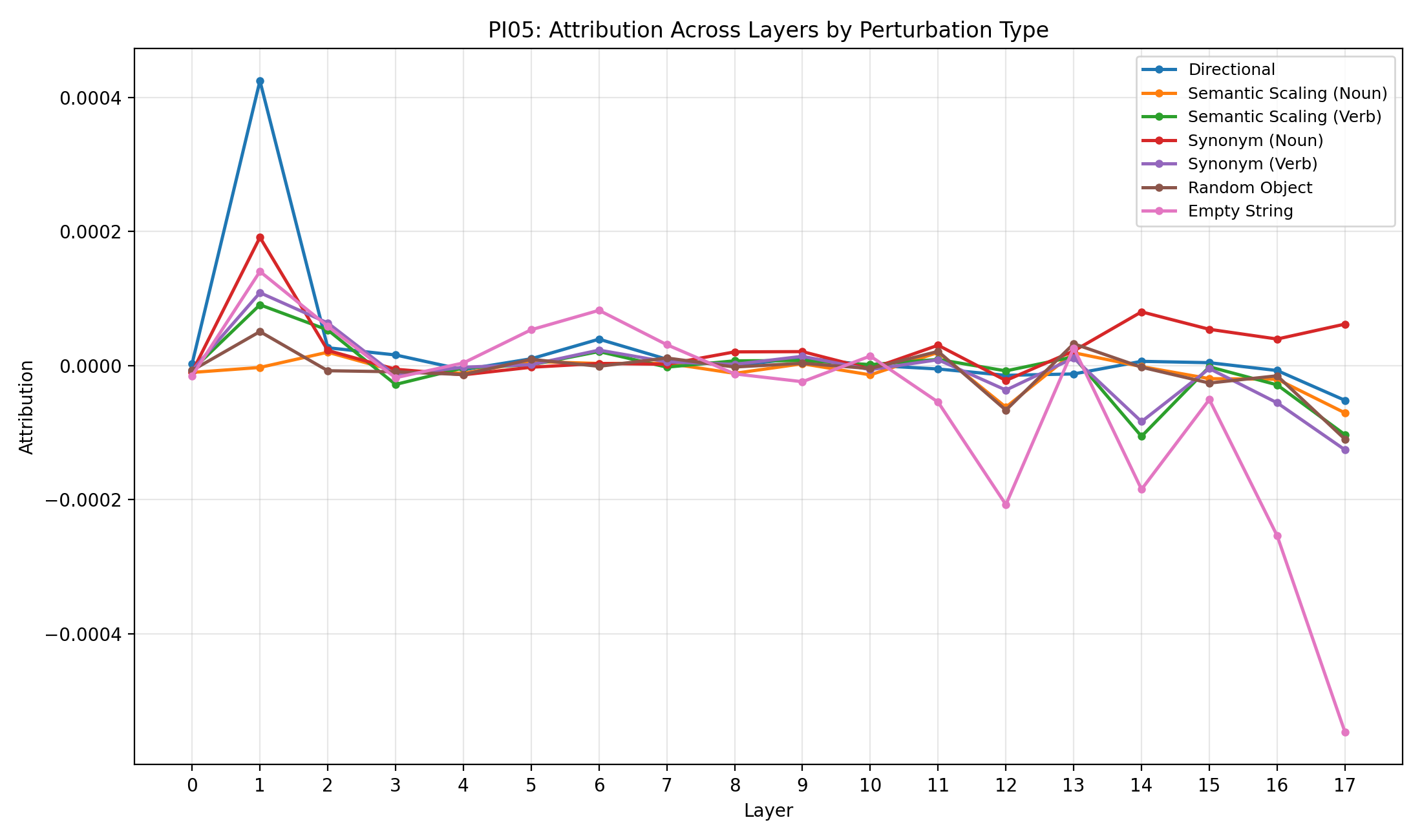}
        \subcaption{$\pi_{0.5}$}
    \end{subfigure}%
    \begin{subfigure}[b]{0.5\textwidth}
        \includegraphics[width=\linewidth]{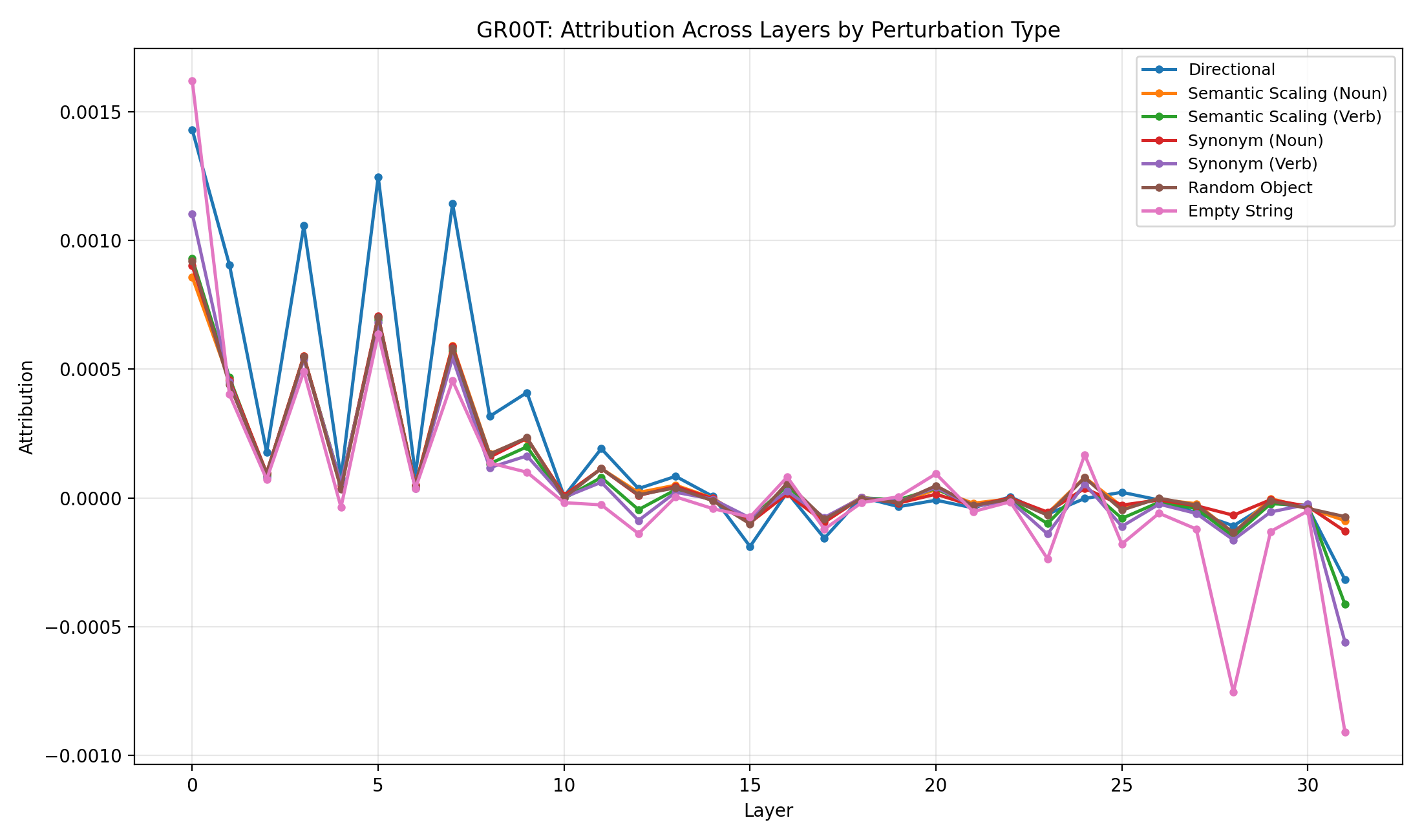}
        \subcaption{GR00T N1.7}
    \end{subfigure}
    \caption{Attribution effect per layer and perturbation type. Positive attribution indicates that the perturbation decreased the metric relative to the clean run, while negative attribution indicates an increase; larger magnitudes reflect a stronger causal effect of that layer on the model's output.}
    \Description{Attribution effect per layer and perturbation type. Positive attribution indicates that the perturbation decreased the metric relative to the clean run, while negative attribution indicates an increase; larger magnitudes reflect a stronger causal effect of that layer on the model's output.}
    \label{fig:att_per_layer_type}
\end{figure*}
\section{Results}
\subsection{Attribution Patching Results}
\textbf{Directional perturbations elicit the largest attribution effects in the early layers of both $\pi_{0.5}$ and GR00T N1.7.}
We report the attribution patching effect per layer in Figure~\ref{fig:att_per_layer_type}. 
In both models, we find that the magnitude and layer-wise distribution of attribution effects differ by linguistic category. 
Directional perturbations elicit substantially larger attribution patching effects than object-referring perturbations.
Particularly, layer 1 for $\pi_{0.5}$ and layers 3, 5, and 7 for GR00T N1.7 stand out as eliciting a strong reaction to directional changes.
In $\pi_{0.5}$, verb perturbations produce slightly stronger attribution effects than noun perturbations, suggesting the action expert is more sensitive to changes that alter the described action (verb, directional) than to changes in the referenced object (noun).
The higher attribution of action-altering perturbations is highlighted especially when replacing nouns with random objects, where referencing non-visible objects affects the model less than perturbations on the verb phrases.

GR00T N1.7 reacts generally strongest to directional perturbations in the early layers. 
In contrast, the effect of other perturbation types remains almost identical in layers 1-7, and mainly diverges in the early middle layers (9-17). 
In late layers, passing an empty string elicits the strongest reaction.
Contrary to $\pi_{0.5}$, GR00T N1.7 is more reactive to changes with respect to perturbations of the noun phrases than to verb phrases. 
The actual existence of an object that the corresponding noun phrase describes is not of much importance, as the random object substitution has almost the same attribution effect as replacing a noun with a synonym or hypernym.

Overall, GR00T N1.7 is more sensitive to linguistic perturbations than $\pi_{0.5}$, as highlighted by the overall magnitude of the attribution effects. 

\textbf{Language conditioning is concentrated in the early layers.}
For $\pi_{0.5}$ across all perturbation types, attribution concentrates most strongly in the very early layers, around layer 6, and toward the final layers of the action expert (past layer 12), while GR00T's language conditioning is concentrated primarily in layers 0-10, with the effect diminishing past the centre of the model.
The importance of GR00T N1.7's cross-attention language conditioning is directly visible as an alternating pattern of elevated attribution across the token sequence following each cross-attention layer, i.e. every 2nd layer offset by one (1, 3, 5, ...). 
GR00T N1.7 applies a masking operation during the cross-attention layers, leaving only every 4th layer (0, 4, 8, ...) tasked with incorporating additional linguistic information from the backbone into the residual stream.
We hypothesise that this masking causes the additional spike in attribution at layer 5, followed by continuous decay toward the later layers, where the effects become more stable. 
We observe that ingested language conditioning affects the self-attention layers more clearly, where the injected information might become elevated. 

\textbf{Late-layer attribution can turn negative.}
For both $\pi_{0.5}$ and GR00T N1.7, a subset of the final layers exhibit negative attribution to language corruption, despite the models' overall reliance on early-layer conditioning. 
A possible explanation could be a self-repair (\enquote{Hydra effect}) dynamic documented in language-model transformers \citep{mcgrathHydraEffectEmergent2023}, whereby downstream components partially compensate for perturbed upstream signals, such that corrupting conditioning information can occasionally align the residual stream more closely with the training-time target than the unperturbed forward pass.

\begin{figure*}
    \centering
    \begin{subfigure}[b]{0.5\textwidth}
        \includegraphics[width=\linewidth]{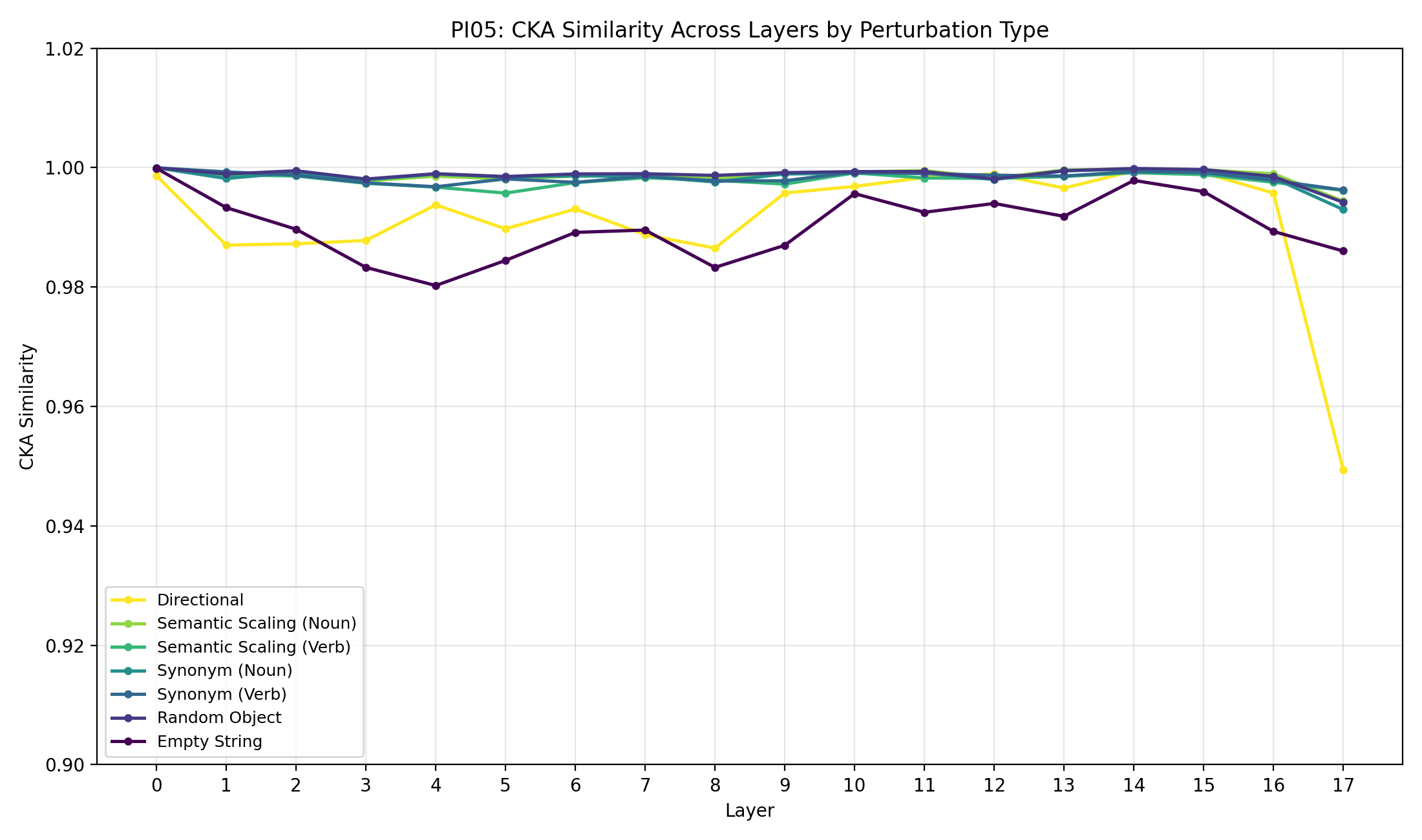}
        \subcaption{$\pi_{0.5}$}
    \end{subfigure}%
    \begin{subfigure}[b]{0.5\textwidth}
        \includegraphics[width=\linewidth]{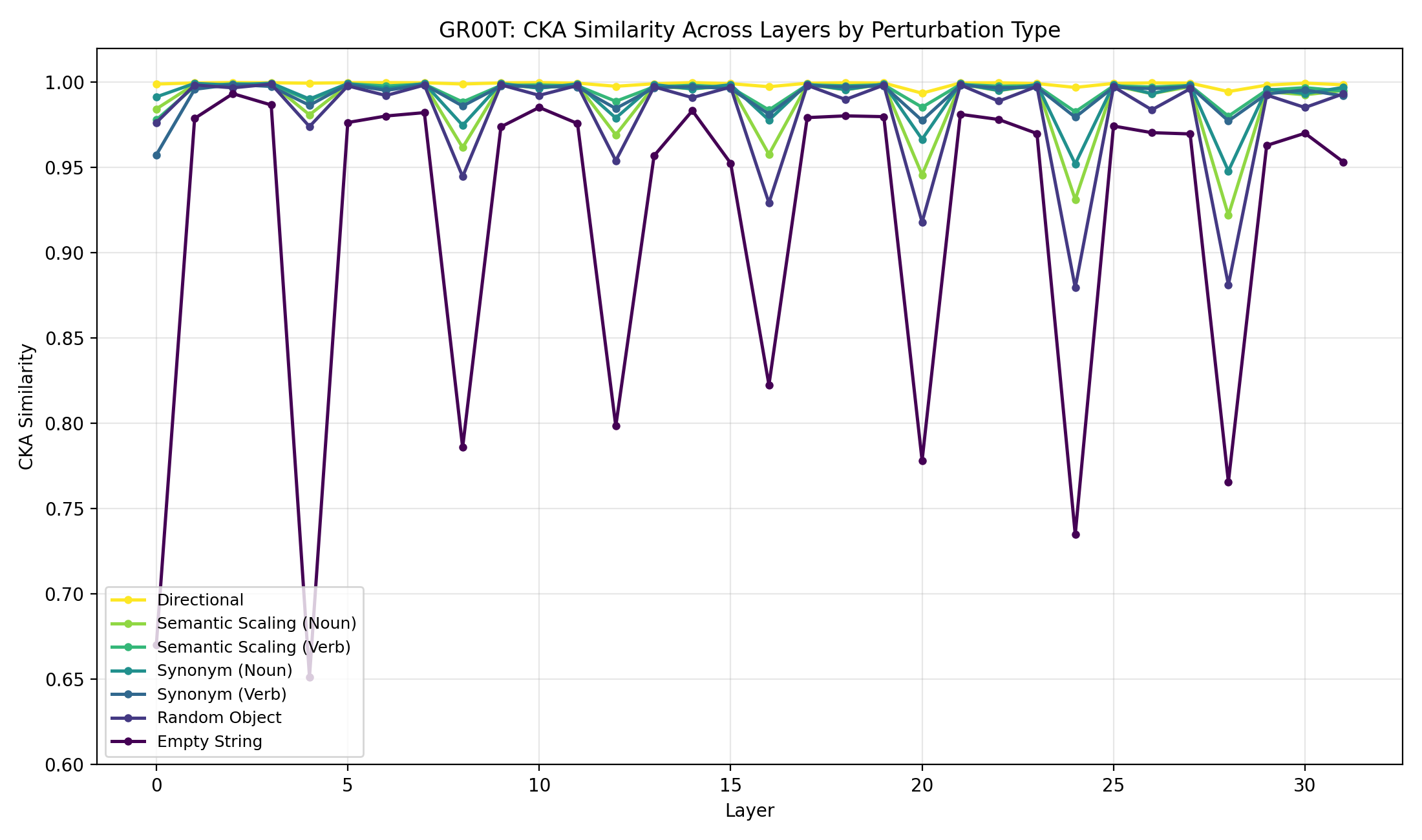}
        \subcaption{GR00T N1.7}
    \end{subfigure}
    \caption{Per-layer CKA between clean and perturbed representations, computed for each perturbation type. Lower CKA indicates greater representational divergence from the clean run.}
    \Description{Per-layer CKA between clean and perturbed representations, computed for each perturbation type. Lower CKA indicates greater representational divergence from the clean run.}
    \label{fig:cka_per_layer}
\end{figure*}

\subsection{Representational Analysis}
\textbf{Embedding space could correspond to semantics.}
We utilise CKA on the residual stream after the attribution patching interventions. 
Figure~\ref{fig:cka_per_layer} highlights the changes in CKA per layer for both VLAs.
We find that $\pi_{0.5}$'s CKA remains overall robust and only elicits visible differences on perturbation types empty string and directional. 
Again, inserting non-existing objects does not affect the model much.
GR00T N1.7 is much more reactive in the embedding space to language perturbations. 
The spiking pattern repeating every 4 layers further highlights the importance of the cross-attention mechanism that incorporates linguistic information. 
While directional perturbations yielded the highest attribution effect, the embedding space remains largely unaffected. 
We hypothesise that this is due to directional modifiers being very close in semantic word embedding spaces due to being the same word type, of similar length and associated with similar contexts. 
Otherwise, specific perturbations change the embedding space much more, closely following our proposed perturbation severity ranking, with synonyms having the least effect and empty strings having the largest effect. 

\begin{figure}[ht]
    \begin{subfigure}[b]{0.4\textwidth}
    \includegraphics[width=\linewidth]{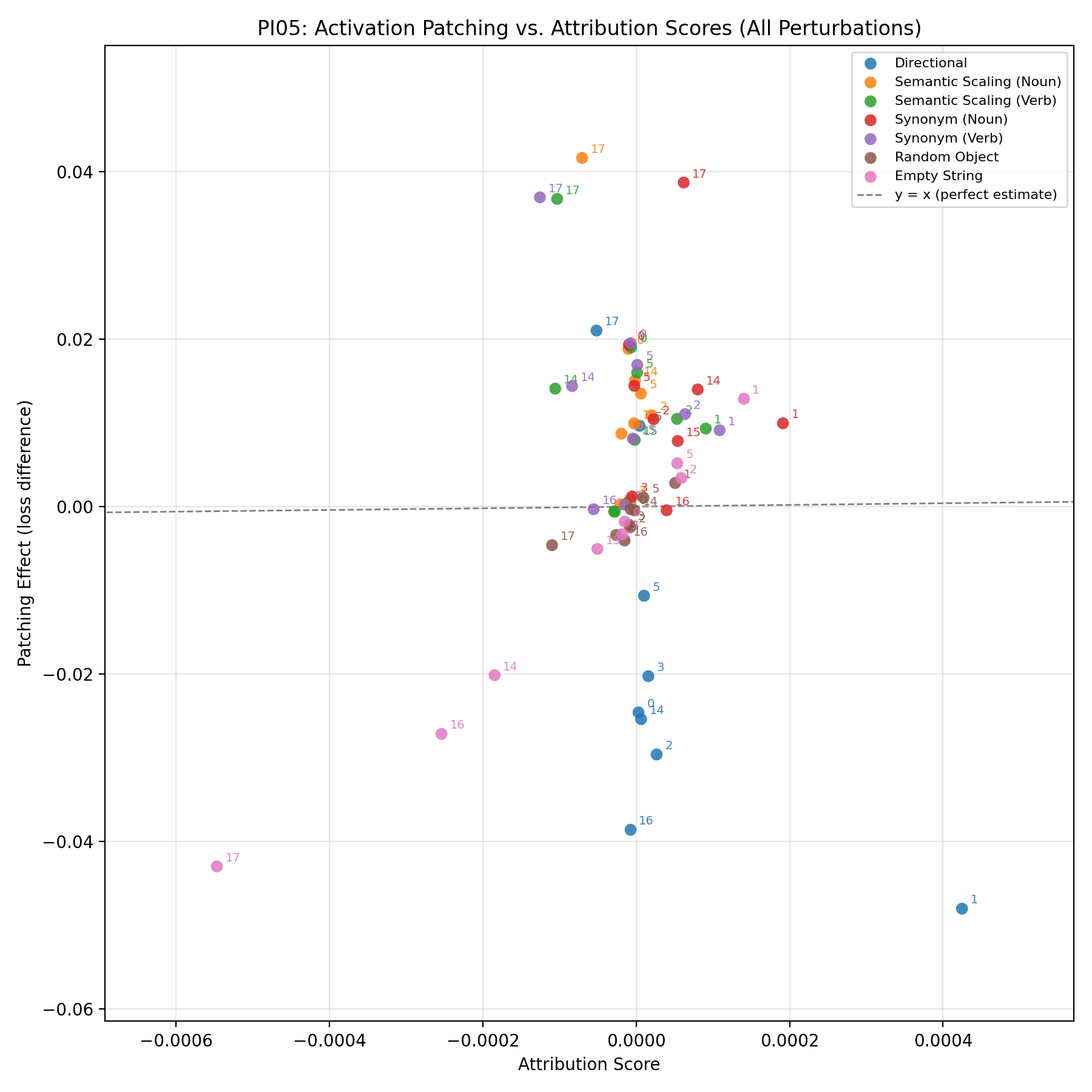}
    \subcaption{$\pi_{0.5}$}
\end{subfigure}
\begin{subfigure}[b]{0.4\textwidth}
    \includegraphics[width=\linewidth]{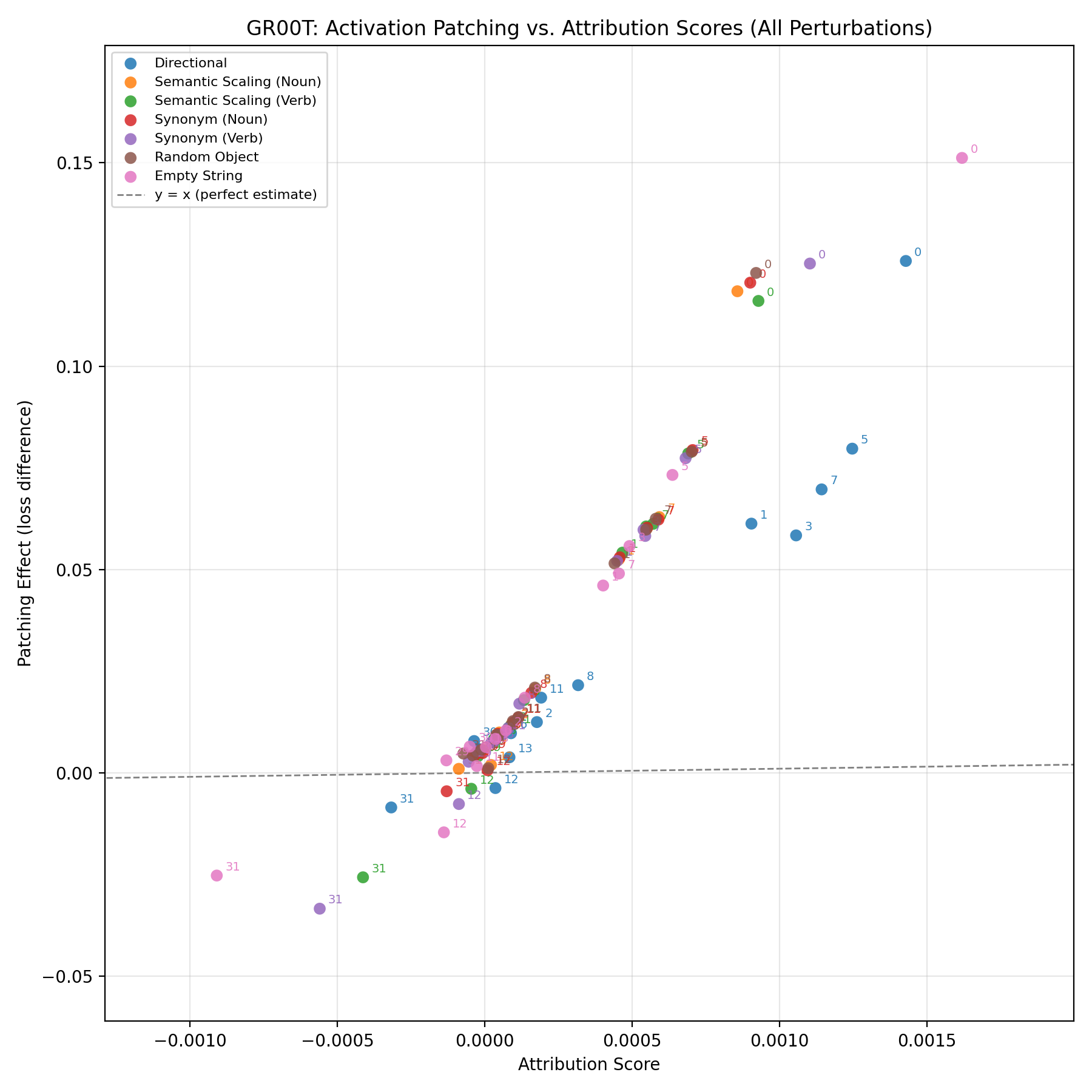}
    \subcaption{GR00T N1.7}
\end{subfigure}
\caption{Activation Patching effect plotted against attribution patching effect}
\Description{Activation Patching effect plotted against attribution patching effect}
\label{fig:atp_vs_acp}
\end{figure}

\textbf{Attribution patching underestimates causal effect relative to activation patching}
We ran additional experiments on a subset of available model layers using activation patching. 
Comparing the attribution patching effect against ground-truth activation patching in Figure~\ref{fig:atp_vs_acp}, we find that attribution patching systematically underestimates the causal effect of most layers in $\pi_{0.5}$. 
This divergence is most pronounced in early layers (e.g., layers 0-1, 5), which are at times substantially less reactive under attribution patching than under activation patching.
This is consistent with the linear (first-order gradient) approximation of attribution patching breaking down further from the output, where non-linear interactions accumulate across more downstream layers.
Additionally, attribution patching underestimates the effects of patching, as shown by the difference in magnitude between the activation patching and the attribution patching effects.
In several layers (2, 3, 5, 14) of $\pi_{0.5}$ under directional perturbation, the discrepancy is more severe than elsewhere: attribution patching (x-axis) estimates a near-zero to slightly positive effect (~0.0002-0.0013), while activation patching (y-axis) finds a strongly negative effect (~-0.05-0.35).
This suggests the directional perturbation induces a qualitatively different, non-linear response at layers (2, 3, 5, 14), potentially reflecting a threshold or saturation effect that only manifests under directional shifts.
However, the trends we observed using attribution patching remain: 
Early layers are most affected by language perturbations, with the effect being prominently visible for GR00T N1.7 (Figure~\ref{fig:atp_vs_acp}b right half).

\subsection{Comparing Activation and Attribution Patching}
To quantify agreement between attribution and activation patching beyond visual inspection, we compute Pearson and Spearman rank correlations between the two sets of effects for each model, and report them in Table~\ref{tab:attr_vs_activation_corr}.
For GR00T N1.7, we observe a strong positive correlation (Pearson $r=0.946$, $p=8.26\times10^{-49}$; Spearman $\rho=0.972$, $p=2.48\times10^{-62}$), indicating that attribution patching closely approximates both the magnitude and rank-ordering of ground-truth activation patching effects. 
In contrast, for $\pi_{0.5}$ we find no significant correlation (Pearson $r=0.089$, $p=0.490$; Spearman $\rho=0.097$, $p=0.448$), suggesting that the linear approximation underlying attribution patching fails to capture the true causal effects in this model. 
This discrepancy highlights that attribution patching's reliability as a proxy for activation patching is model-dependent and should not be assumed to generalise across architectures.

\begin{figure}[t]
    \centering
    \begin{subfigure}[b]{0.45\textwidth}
        \includegraphics[width=\linewidth]{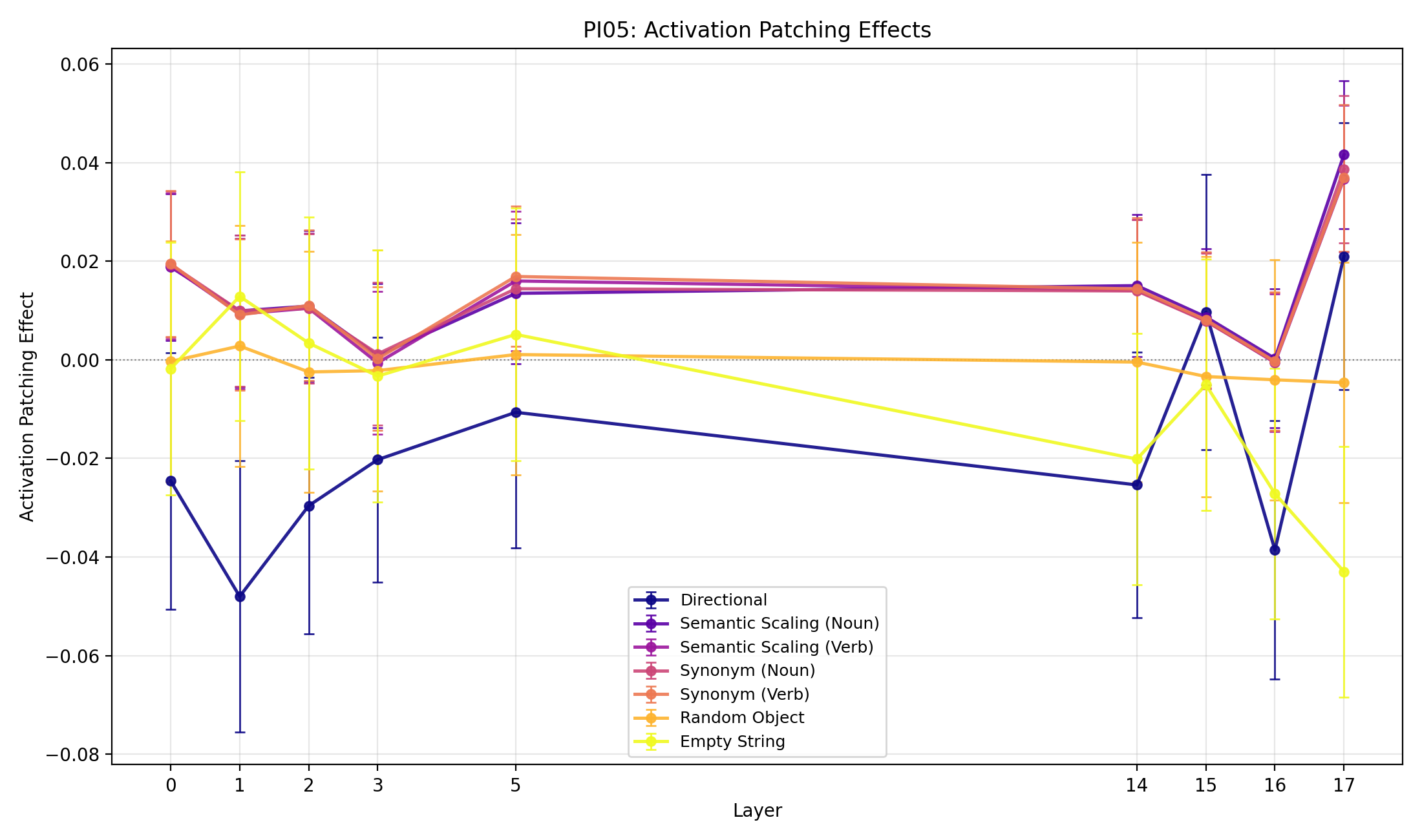}
        \subcaption{$\pi_{0.5}$}
    \end{subfigure}
    \begin{subfigure}[b]{0.45\textwidth}
        \includegraphics[width=\linewidth]{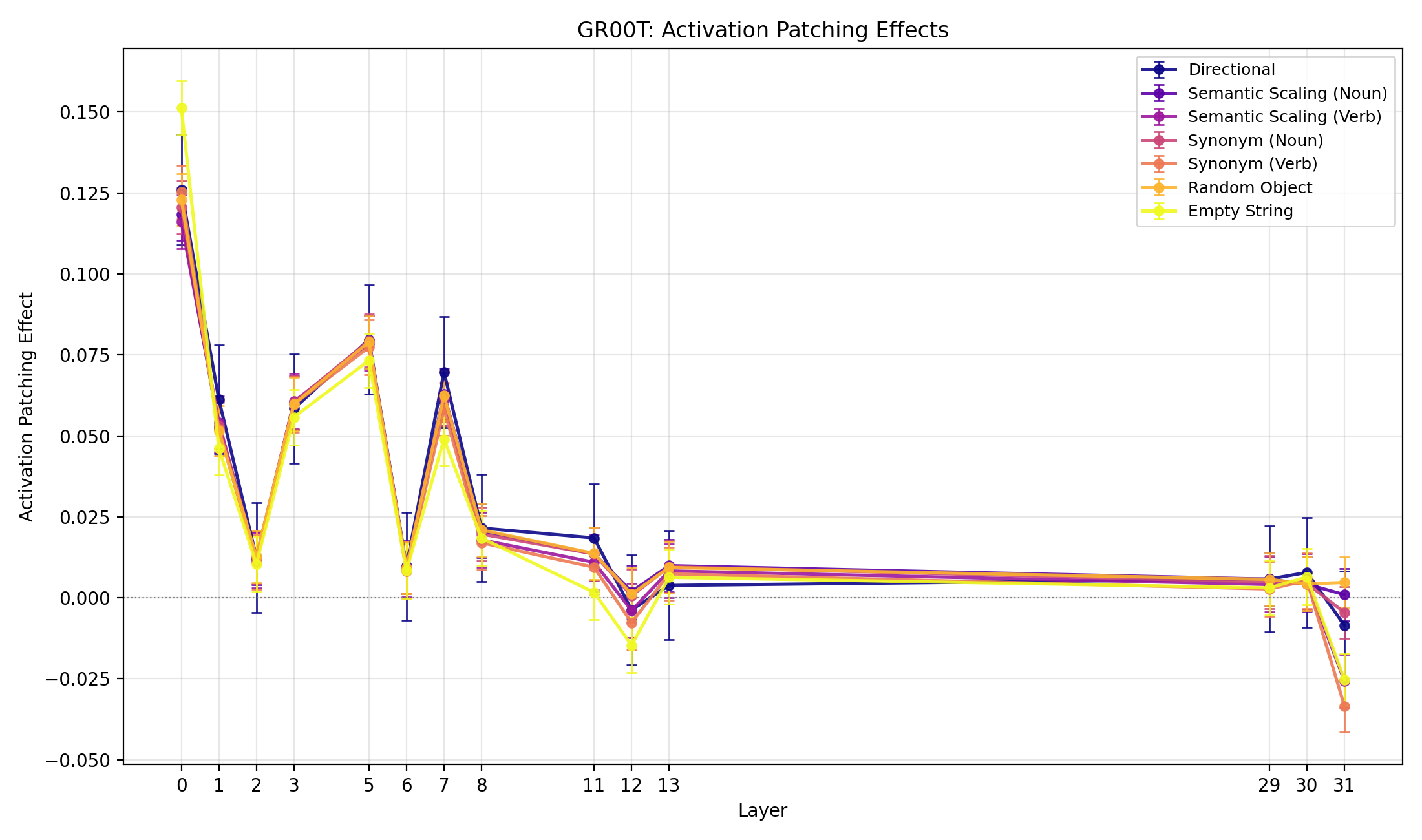}
        \subcaption{GR00T N1.7}
    \end{subfigure}
    \caption{Per-layer activation patching effect of selected layers.}
    \Description{Per-layer activation patching effect of selected layers.}
    \label{fig:act_patch_per_layer}
\end{figure}

\begin{table}[t]
\centering
\caption{Correlation between attribution patching and activation patching effects.}
\label{tab:attr_vs_activation_corr}
\begin{tabular}{lcc}
\toprule
Model & Pearson $r$ & Pearson $p$ \\
\midrule
GR00T N1.7  & 0.946 & $8.26 \times 10^{-49}$  \\
$\pi_{0.5}$ & 0.089 & $4.90 \times 10^{-1}$   \\
$\pi_{0.5}$ w/o Directional & 0.097 & $4.86 \times 10^{-1}$  \\
\midrule
Model &  Spearman $\rho$ & Spearman $p$ \\
\midrule
GR00T N1.7  & 0.972 & $2.48 \times 10^{-62}$ \\
$\pi_{0.5}$ & 0.097 & $4.48 \times 10^{-1}$  \\
$\pi_{0.5}$ w/o Directional & 0.2508 & $6.73 \times 10^{-2}$  \\
\bottomrule
\end{tabular}
\end{table}

Figure~\ref{fig:act_patch_per_layer} displays the activation patching effects of selected layers in both VLAs. 
Activation patching follows a trajectory broadly similar to the reported attribution patching effects. 
For $\pi_{0.5}$, however, a notable discrepancy arises for directional perturbations: activation patching yields a negative effect, whereas attribution patching predicts the largest positive effect among all perturbation types. 
For the remaining perturbations, attribution patching tracks activation patching reasonably well in the early layers but increasingly underestimates the effect magnitude in later layers.

Excluding directional perturbations improves the correlation between the two methods (Pearson $r=0.097$, $p=4.86\times10^{-1}$; Spearman $\rho=0.251$, $p=6.73\times10^{-2}$). 
The Spearman correlation increases roughly 2.5-fold with a corresponding drop in $p$-value, suggesting improved agreement in the rank ordering of layer effects, while the near-unchanged Pearson correlation indicates that attribution patching still fails to capture the correct magnitude of activation patching effects for $\pi_{0.5}$.
\section{Discussion}
Our finding that $\pi_{0.5}$'s internal representations remain stable under different perturbation types, including null prompts, is in accordance with \citet{grantNotAllFeatures2026}, who report that fine-tuned VLA policies are effectively vision-only, with injected visual activations alone recovering near-identical behaviour under null prompts. 
We see our results as confirmatory while digging deeper into where language is actually processed and extending beyond the null prompt to more fine-grained perturbations.
As such, \citeauthor{grantNotAllFeatures2026}'s intervention (null-prompt / cross-task activation injection) probes whether language is \emph{necessary} for the specific behaviours they measure at the level of task success, whereas our perturbations probe whether altering language content \emph{changes} the internal computation and its causal footprint on the predicted action, without requiring that this sensitivity be sufficient to flip task success outcomes.
A model can be internally sensitive to language perturbation while still substantially recovering behaviour from vision alone when language is entirely absent. 
Second, this distinction mirrors the attribution effect/CKA dissociation we observe when comparing Figures~\ref{fig:att_per_layer_type} and~\ref{fig:cka_per_layer}: representational change under perturbation does not imply that the change is behaviourally load-bearing.
Additionally, our activation-patching verification indicates that even where the attribution patching effect is small, the true causal effect can be larger.  
Fully reconciling representational sensitivity with behavioural outcome, however, would require causal verification that neither study completes on its own.
We view our finding and theirs as complementary: language pathways may contribute little under coarse manipulations like theirs, yet remain mechanistically active in ways only fine-grained, meaning-preserving perturbations like ours reveal.

Regarding COAST~\citep{miaoContrastiveConceptorActivation2026}, our results are complementary, too: 
COAST finds that VLA policies retain substantial task-relevant knowledge in low-rank latent subspaces that outcome-driven steering can unlock, implying that these models often encode more task-relevant information than they act on by default. 
Again, this resonates with our own layer-wise dissociations between representational change and causal effect (Figures~\ref{fig:att_per_layer_type} and~\ref{fig:cka_per_layer}). 
In both cases, information present in the residual stream is not fully reflected in the model's default behaviour. 
COAST demonstrates this at the level of task success/failure subspaces via steering, while we demonstrate it at the level of layer-wise attribution under targeted language perturbation. 

\section{Future Work}
Our study suggests several directions for future work.
For instance, our perturbations are synthetic (directional swaps, synonym and semantic-scale substitution) and do not capture real-world variation such as ambiguity or compositional phrasing. 
LIBERO instructions are also short, templated, and object-centric, and their constrained scene layouts may let visual context compensate for linguistic ambiguity, likely understating language sensitivity in less visually redundant settings. 
Extending the analysis to benchmarks with richer instruction diversity or real-world deployment logs would help establish the generalisability of our findings. 
Since all perturbed predictions are evaluated against the original demonstration, our results show that language influences the action, so closed-loop rollouts or counterfactual targets for meaning-changing perturbations would test whether the observed sensitivity reflects appropriate use of the instruction. 
Additionally, applying our perturbation framework to vision would test whether the vision pathway shows analogous causal/representational dissociations.

\section{Conclusion}
We investigated how VLAs ground actions in language using a combined causal and representational pipeline: attribution patching, verified against ground-truth activation patching, alongside CKA to track representational change. 
Applied to $\pi_{0.5}$ and GR00T N1.7, we find language conditioning is model-specific in both sensitivity and locus, and that causal and representational effects can dissociate.
We observe this most clearly for directional language in GR00T N1.7, which drives large attribution effects with little representational change.
Additionally, we find that the reliability of attribution patching is model-dependent: it closely tracks activation patching for GR00T N1.7 but not for $\pi_{0.5}$, especially in early layers.
These findings offer a methodological caution for interpretability work on VLAs and evidence that action experts depend on language non-uniformly, shaped by their conditioning architecture.
\bibliographystyle{unsrtnat}
\bibliography{references}  

\appendix

\section{Implementation Details}
We use \texttt{nnsight} to batch clean and corrupted runs within a single tracing context.
A model wrapper encapsulates all architecture-specific details (layer paths, activation
targets, metric definition), keeping the tracing loop model-agnostic.
To ensure comparability across experiments, we initialise all randomisers with $seed=42$. 
This also ensures that our LIBERO-10 subset of 10,000 samples remains comparable across runs. 
Code will be released.



\section{Random Object List}
Listing~\ref{listing:random_words} displays the full list of objects for substituting the nouns in the task instruction with a non-present object.
\lstset{
    basicstyle=\ttfamily\footnotesize,
    breaklines=true,
    breakatwhitespace=true,
    breakindent=0pt,          
    breakautoindent=false,    
    numbers=none,             
    columns=fullflexible,
    frame=tb,
    backgroundcolor=\color{gray!8},
    rulecolor=\color{gray!60},
    xleftmargin=3pt,
    xrightmargin=3pt,
    framexleftmargin=3pt,     
    framexrightmargin=3pt
}
\begin{lstlisting}[
    float=h,
    caption={Random object list},
    label={listing:random_words}
]
toaster, umbrella, bicycle, lantern, spatula, anchor, trombone, cactus, typewriter, kayak, chandelier, harmonica, wheelbarrow, telescope, pillow, compass, violin, helmet, briefcase, shovel, kettle, backpack, microscope, hammock, flashlight, bathtub, clarinet, pinecone, suitcase, sundial, mailbox, accordion, ladder, thermos, pretzel, windmill, snowglobe, stapler, canoe, chalkboard, pocketwatch, birdcage, wrench, trampoline, candlestick, binoculars, wagon, tambourine, scarecrow, periscope
\end{lstlisting}

\clearpage
\section{Synonyms and Hypernyms}
Table~\ref{tab:synonym_hypernym} provides the full list of configured synonyms and hypernyms used to perturb the instructions in the LIBERO-10 benchmark. 
When multiple synonyms are available, we randomly select one of them; when no synonym is available, the word remains unchanged.

\begin{table}[tb]
\small
\centering
\caption{Word list with synonyms and hypernyms}
\label{tab:synonym_hypernym}
\begin{tabularx}{\columnwidth}{lXl}
\toprule
\textbf{Word} & \textbf{Synonyms} & \textbf{Hypernym} \\
\midrule
\multicolumn{3}{l}{\textbf{Nouns}} \\
\midrule
mug & cup, jug, tankard & drinking vessel \\
plate & platter & flatware \\
chocolate pudding & chocolate mousse, chocolate custard & chocolate dessert \\
microwave & - & kitchen appliance \\
stove & cooker, boiler, kiln, heater & kitchen appliance \\
moka pot & stovetop coffee maker & kitchen appliance \\
alphabet soup & letter soup, pasta soup & dish \\
tomato sauce & marinara, passata & pasta sauce \\
cream cheese & soft cheese, spreadable cheese, cheese spread & cheese \\
basket & hamper, crate & container \\
butter & dairy spread & dairy product \\
bowl & dish, basin & vessel \\
drawer & cabinet, compartment & storage space \\
cabinet & case, commode & furniture \\
book & volume, tome & publication \\
compartment & cell, slot & space \\
caddy & canister, crate & can \\
\midrule
\multicolumn{3}{l}{\textbf{Verbs}} \\
\midrule
put & place, position & move \\
place & position, put, set & move \\
close & shut & secure \\
pick up & lift up, gather up & elevate \\
\bottomrule
\end{tabularx}
\end{table}


\end{document}